\documentclass{article}

\usepackage[preprint]{neurips_2024}

\usepackage[utf8]{inputenc} 
\usepackage[T1]{fontenc}    
\usepackage{hyperref}       
\usepackage{url}            
\usepackage{booktabs}       
\usepackage{amsfonts}       
\usepackage{nicefrac}       
\usepackage{microtype}      
\usepackage{xcolor}         
\usepackage{subcaption}

\usepackage{microtype}
\usepackage{graphicx}
\usepackage{booktabs} 

\usepackage{hyperref}       
\usepackage{url}            
\usepackage{amsfonts}       
\usepackage{amsmath}        
\usepackage{nicefrac}       
\usepackage{microtype}      
\usepackage[dvipsnames]{xcolor}
\usepackage{xspace}

\newtheorem{theorem}{Theorem}[section]
\newtheorem{proposition}[theorem]{Proposition}

\newtheorem{corollary}[theorem]{Corollary}

\newtheorem{assumption}[theorem]{Assumption}

\DeclareMathOperator*{\argmax}{arg\,max}

\usepackage{algorithm}
\usepackage{algorithmic}
\newcommand{\modelname}{\texttt{XShare\xspace}}

\title{
XShare: Collaborative in-Batch Expert Sharing for~Faster MoE Inference}

\author{%
 Daniil Vankov$^{1}$\thanks{Work is done during an internship at Amazon.} \quad Nikita Ivkin$^2$ \quad Kyle Ulrich$^2$  \And \vspace{2pt} Xiang Song$^2$ \quad Ashish Khetan$^2$ \quad George Karypis$^2$ \vspace{0.5cm}\\
$^1$ASU. \quad $^2$Amazon Web Services \\
\texttt{dvankov@asu.edu}, \texttt{\{ivkin,urlichkr,xiangsx, khetan, gkarypis\}@amazon.com}
}

\begin{document}

\maketitle

\begin{abstract}
Mixture-of-Experts (MoE) architectures are increasingly used to efficiently scale large language models. However, in production inference, request batching and speculative decoding significantly amplify expert activation, eroding these efficiency benefits. We address this issue by modeling batch-aware expert selection as a modular optimization problem and designing efficient greedy algorithms for different deployment settings. The proposed method, namely \modelname, requires no retraining and dynamically adapts to each batch by maximizing the total gating score of selected experts. It reduces expert activation by up to 30\% under standard batching, cuts peak GPU load by up to 3x in expert-parallel deployments, and achieves up to 14\% throughput gains in speculative decoding via hierarchical, correlation-aware expert selection  even if requests in a batch drawn from heterogeneous datasets.

\end{abstract}

\section{Introduction}
\label{sec:intro}
Mixture-of-Experts (MoE) architectures have emerged as a powerful paradigm for scaling neural networks to unprecedented sizes while maintaining computational efficiency during training. The key insight, introduced by \citet{Shazeer2017}, is to replace feed-forward layers with a collection of fully connected networks, where only a sparse subset is activated for each input token. This conditional computation approach fundamentally changes the scaling characteristics of neural networks: by decoupling total model capacity (parameters) from per-token computation (FLOPs), MoE models achieve dramatically better performance-to-FLOPs ratios compared to dense models. Recent work on scaling laws for MoE models \citep{abnar2025parameters} demonstrates that for a fixed computational budget, sparse MoE models with 10-100$\times$ more parameters than their dense counterparts can achieve superior accuracy while maintaining similar training costs. For instance, Switch Transformers \citep{Fedus2022} scale to trillion-parameter models while requiring only a fraction of the FLOPs needed to train equivalent-quality dense models. This favorable parameter-to-FLOPs trade-off during training has made MoE the architecture of choice for state-of-the-art large language models, with architectures like GShard \citep{Lepikhin2020} successfully demonstrating that sparse expert activation enables efficient training of models at unprecedented scales.

However, the training-time efficiency gains of MoE architectures do not directly translate to inference-time efficiency in production serving systems. Production LLM serving systems can be categorized by their optimization objective: \emph{latency-optimized} systems minimize per-request latency (typically using batch size 1), \emph{throughput-optimized} systems maximize requests per second, and \emph{goodput-optimized} systems maximize throughput subject to service level objectives for latency. Our work primarily targets goodput-optimized deployments, which must balance batching for hardware efficiency against latency constraints. These systems introduce two key challenges that make MoE inference inefficient. First, to maximize hardware utilization and amortize the cost of memory transfers, serving systems batch multiple requests together. While each individual token activates only $k$ experts (typically from $k=1$, $k=4$, or $k=8$), the union of experts activated across all tokens in a batch grows substantially with batch size. Second, virtually all modern serving systems employ speculative decoding to improve throughput, where a small draft model generates candidate tokens that are verified in parallel by the target model. For a speculative length of $s$, each request effectively multiplies the batch size by $(s+1)$: a batch of $B$ requests becomes $B(s+1)$ tokens that must be processed simultaneously. The combination of request batching and speculative decoding results in expert activation patterns, where, for moderate batch sizes, the majority of experts must be loaded into memory, and for large batches, nearly all experts are activated. 
To quantify this effect, the expected number of activated experts in a batch follows
\[\mathbb{E}[N_a] = N\left(1-(1-k/N)^B\right),\]
where $N$ is the total number of experts, $k$ is the top-$k$ routing, and $B$ is the effective batch size (including speculative tokens). For DeepSeek-R1 ($N=256$, $k=8$), this predicts that at batch size 8, approximately 57 experts are activated, and at batch size 32, approximately 163 experts—closely matching our empirical observations. With speculative decoding of length $s$, the effective batch becomes $B(s+1)$, further accelerating expert activation growth. 
While a sparse MoE model remains computationally more efficient than a dense model with equivalent parameters (only activated experts consume FLOPs), the memory bandwidth bottleneck dominates during the autoregressive decode phase: all activated experts must be loaded from memory regardless of their utilization, making inference memory-IO-bound rather than compute-bound.

\begin{figure}[H]
  \centering
    \centering
    \includegraphics[width=0.6\textwidth]{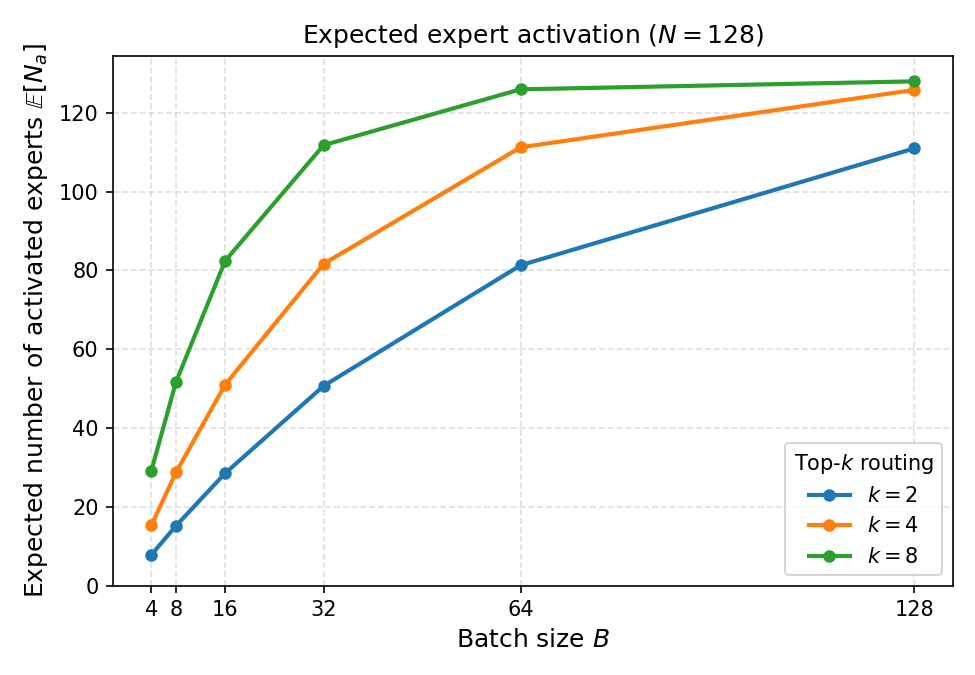}
  \caption{Average number of activated experts}
\end{figure}

Recent interpretability work on MoE models \citep{Ying2025beyond} demonstrates that stronger models achieve better performance with lower neuron utilization — suggesting that efficient sparse activation is a hallmark of model quality rather than a compromise. However, these insights have not been translated into practical inference optimization strategies that can dynamically adapt expert selection to batch characteristics.

\textbf{Related works.} Traditional network pruning techniques such as MetaPruning \citep{Liu2019} and BilevelPruning \citep{Gao2024} have been adapted to MoE architectures. More recently, \citet{liu2024efficient} demonstrates that not all experts contribute equally to model accuracy, enabling aggressive static pruning of less important experts. DiEP \citep{Bai2025} advances this direction by learning non-uniform pruning ratios across layers. \citet{xie2024pruner} proposes MoE-Pruner, which uses router-weighted pruning metrics for one-shot weight-level pruning on calibration data, but produces static pruning decisions dependent on the calibration datasets and requires knowledge distillation to recover from moderate accuracy degradation. While these methods reduce the total parameter count, they require retraining or fine-tuning and produce fixed, static pruning decisions that cannot adapt to different workload characteristics or speculative decoding patterns.

Most closely related to our work, \citet{lu2024not} proposes Dynamic Skipping method, which operates at the token level, skipping expert $e_1$ if $w_{e_1} < \beta w_{e_0}$, where $\beta$ is calibrated per-layer. 
Intuitively, an expert is skipped when there is a large diminishing return, reflected by a sharp drop in gating scores between consecutively ranked experts.
However, it processes each token independently and does not account for batch composition, so the batch-level expert explosion problem in production remains unresolved. 

In contrast, Lynx \citep{Gupta2024} was the first to recognize the batch-level expert explosion problem and to propose a batch-aware expert selection policy, LYNX-Lat.
LYNX-Lat aggregates per-token expert preferences across a batch to identify the least-used experts, then drops a fixed, pre-selected number of them. The number of experts dropped is determined offline by tuning the accuracy–performance tradeoff.  However, dropping the least frequently used experts ignores how important those experts were when they were selected (e.g., their rank for the tokens that used them). As a result, it can remove highly ranked experts for some tokens and severely hurt accuracy. Authors describe this policy at a conceptual level, without providing precise algorithms or pseudocode, making the approach difficult to reproduce or directly compare against. 

During the preparation of this manuscript, we became aware of concurrent work by \citet{Oncescu2025Opportunistic} where authors propose building a shared pool of candidate experts by taking the top-$k'$ experts (with $k'<k$) for each token, then filling the remaining  $(k-k')$  slots by reusing experts from that pool — i.e. piggybacking on experts already selected by other tokens in the batch. The idea of creating a shared pool of top $k'$ is similar to Warm-up routine which we implemented in Algorithm \ref{algo:greedy-full}. However, \citet{Oncescu2025Opportunistic} selected the experts for the shared pool based on individual token preferences, so its usefulness for other tokens is often limited, thus can hurt the accuracy.

\begin{figure}[h]
  \centering
    \centering
    \includegraphics[width=\linewidth]{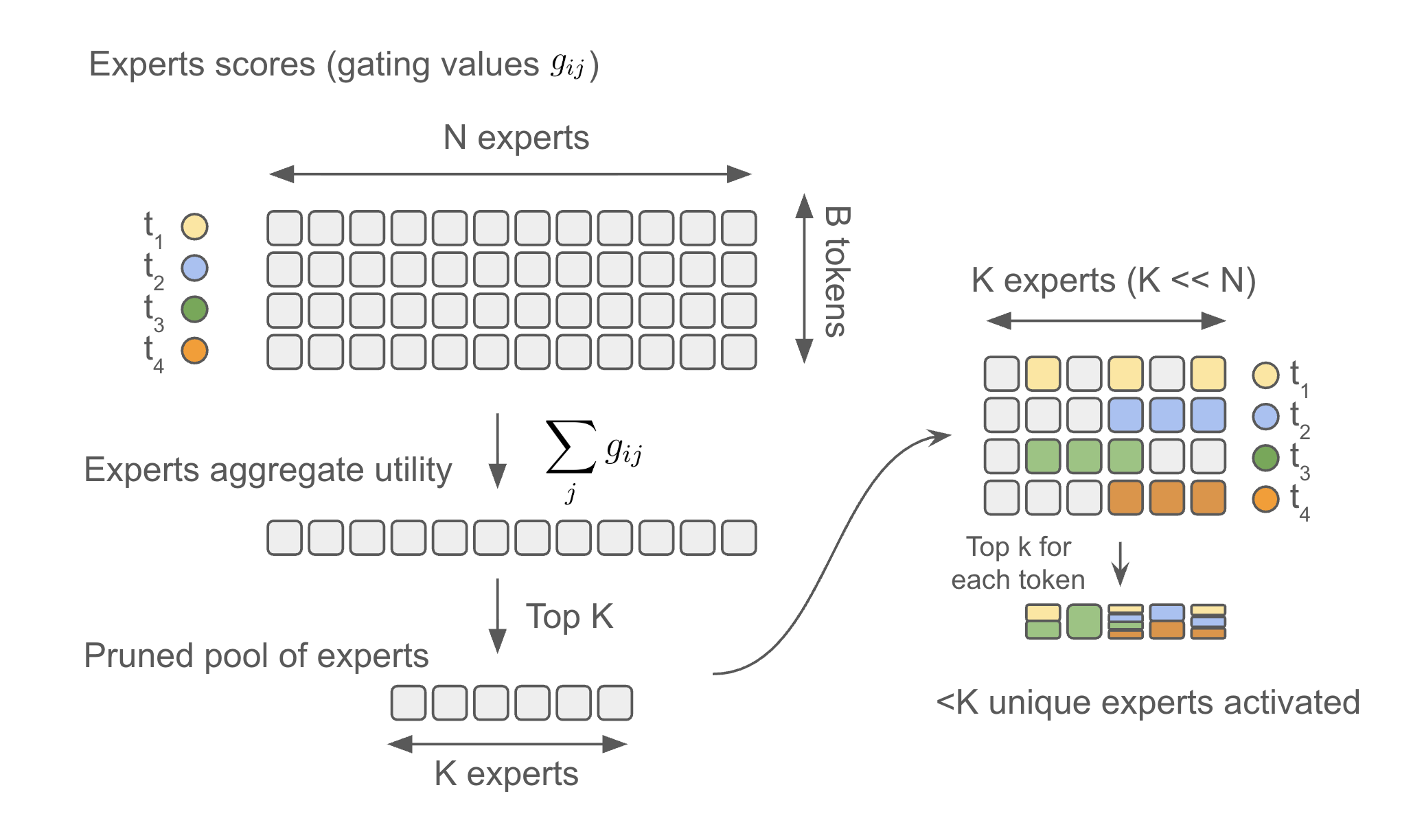}
  \caption{Batch utility expert pruning}
  \label{fig:intuitive-diagram}
\end{figure}

\textbf{Core idea.} Our method is motivated by the observation that expert selection should be optimized for the batch as a whole rather than on a per-token basis. We first aggregate gating scores across all tokens and select the top-$K$ experts according to this batch-level utility. 
This selected subset then constrains per-token routing, from which each token chooses its own top-$k$ experts as it depicted on the Figure \ref{fig:intuitive-diagram}. Our approach further extends to production-critical settings: expert-parallel deployments across GPUs and speculative decoding,
which existing methods do not address.

\textbf{Our Contributions.} In this paper, we mathematically formulate the batch-aware expert selection problem, reduce it to a modular optimization problem, and propose efficient inference-time pruning methods tailored to different deployment scenarios:
\begin{enumerate}
    \item We formulate expert selection as maximizing a modular proxy objective (sum of gating scores) under cardinality constraints, enabling efficient greedy optimization with theoretical guarantees.
    \item We develop a practical three-step algorithm (warm-up, greedy optimization, refinement) that requires no retraining and adapts dynamically to each batch. It reduces number of activated experts by up to 30\% and improves end-to-end OTPS by up to 13\%, while maintaining accuracy within 1\% drop. 
    \item In speculative decoding scenarios, we demonstrate and exploit strong expert-preference correlation among speculative tokens via novel hierarchical expert selection approach, providing speedups up to 14\%.  We show that our method remains robust under mixed-dataset evaluation, where requests originate from different datasets.
    \item For expert parallel (EP) deployments, we propose an EP-aware greedy algorithm that minimizes the maximum per-GPU load. It reduces peak GPU load from 25.6 to 8.6 experts (3$\times$ improvement) while preserving the accuracy.
\end{enumerate}

\section{Preliminaries}
We consider a pre-trained Large Language Model (LLM) incorporating Mixture-of-Experts (MoE) layers. 
This section formalizes the MoE layer operation during the inference phase, assuming all model parameters are fixed.

\subsection{MoE Layer Definition}
Let $x \in \mathbb{R}^d$ denote the input hidden state corresponding to a specific token, where $d$ is the model dimension. An MoE layer consists of a set of $N$ routing expert networks $\{E_i\}_{i=1}^N$ and $N_s$ shared expert networks $\{{E^s_i(x)}\}_{i=1}^{N_s}$, where each expert $E_i: \mathbb{R}^d \to \mathbb{R}^d$ is typically a feed-forward neural network (FFN).

The output of an MoE layer, $y \in \mathbb{R}^d$, is a weighted sum of outputs of experts, mediated by a gating network. Formally, the output is defined as:
\begin{equation*}
    y = \sum_{i=1}^N g_i(x) \cdot E_i(x) + \sum_{j=1}^{N_s} E^s_j(x)
\end{equation*}
where $g(x) \in \mathbb{R}^N$ is the output of the gating function, and $g_i(x)$ represents the scalar gating value (or confidence score) assigned to expert $i$ for input $x$.

\subsection{Gating and Routing Mechanism}
The gating network determines the routing strategy, i.e. which experts are activated and how their outputs are weighted. In modern sparse MoE architectures (e.g., GShard, Switch Transformer, Mixtral), a Top-$k$ routing mechanism is employed to enforce sparsity.

Let $W_g \in \mathbb{R}^{N \times d}$ be the trainable weight matrix of the router. It first computes the raw routing logits $h(x) = W_g x$. To ensure sparse activation, it selects the top $k$ indices with the highest logits, denoted by the set $\mathcal{T}$. The gating values $g(x)$ are computed as follows:
\begin{equation*}
    \begin{aligned}
    g_i(x) =& 
    \begin{cases} 
    \bar{g}_i(x)  & \text{if }  g_i(x)  \in {\rm{Top}} k (g(x))\\
    0 & \text{otherwise}
    \end{cases}, \cr  \bar{g}_i(x) =& \frac{\exp(h_i(x))}{\sum_{j \in \mathcal{T}} \exp(h_j(x))}.
    \end{aligned}
\end{equation*}
Here, $k \ll N$ (typically from $k=1$, $k=4$, or $k=8$), ensuring that the majority of $g_i(x)$ are zeros. During inference, this sparsity allows the model to bypass the computation of $E_i(x)$ for all $i \notin \mathcal{T}$, significantly reducing Floating Point Operations (FLOPs) while maintaining high model capacity.

\subsection{Inference Setting}
In the inference setting, we assume the parameters of the experts $\{E_i\}$ and the gating weights $W_g$ are frozen. The problem of efficient MoE inference reduces to identifying the set $\mathcal{T}$ and \emph{retrieving the specific expert weights for computation}.

\textbf{Decoding Phase Focus.} Our methods target the autoregressive \emph{decoding phase} of LLM inference, where the model generates output tokens one at a time. During this phase, the model operates in a memory-bandwidth-bound regime, making expert activation patterns particularly critical. In contrast, the prefill (prompt processing) phase typically processes a single long sequence and already exhibits high expert activation due to the diversity of tokens in the prompt. The decoding phase, however, processes multiple independent sequences in a batch, each generating one token at a time, which is where batching and speculative decoding dramatically increase expert activation and where our optimization provides the most benefit.

\section{Problem Formulation}
\label{sec:problem-formulation}

In this section, we formalize the \emph{batch-aware expert selection problem} and develop an efficient algorithm with theoretical guarantees.

\subsection{Batch-Level Expert Activation}

Consider a batch $B = \{x_1, \ldots, x_n\}$ of $n$ tokens arriving at the model during the decode phase. At each MoE layer $l~\in~\{1, \ldots, L\}$, for each token $x_i$, the router produces scores $G_i^{(l)} = (g_{i,1}^{(l)}, \ldots, g_{i,N}^{(l)})$ over the $N$ experts, where $g_{i,j}^{(l)}~=~g_j^{(l)}(x_i)$ denotes the gating score for expert $j$ at layer $l$. Under standard top-$k$ token routing, each token activates its $k$ highest-scoring experts per layer. However, across the batch, the \emph{union} of activated experts can be much larger than $k$:
\[
\left| \bigcup_{i=1}^{n} \text{TopK}(G_i^{(l)}) \right| \gg k \quad \text{for moderate } n.
\]
As shown in Section~\ref{sec:intro}, this union grows to 62\% and 95\% of all experts for batch sizes 32 and 64, respectively. 

Although the model remains computationally sparse (each token still activates only $k$ experts), the large fraction of activated experts must be loaded into memory, creating a memory-IO bottleneck that dominates inference latency during the decode phase.

\subsection{Optimization Objective}

Our goal is to select small subsets of experts $S_l \subseteq E$ for each layer $l$ such that routing each token to its top-$k$ experts \emph{within} $S_l$ preserves model accuracy, while keeping the total number of activated experts across all layers small. Formally, given a batch $B$, we seek:
\begin{equation}
\begin{aligned}
\label{eq:original-problem}
\max_{S_1, \ldots, S_L \subseteq E} \quad & \text{Accuracy}\bigl(f(S_1, \ldots, S_L; B)\bigr) \\
\text{s.t.} \quad & \sum_{l=1}^{L} |S_l| \leq K,
\end{aligned}
\end{equation}
where $f(S_1, \ldots, S_L; B)$ denotes the model output when layer $l$ is restricted to expert set $S_l$, and $K \ll L \cdot N$ is the global expert budget across all layers.

Problem~\eqref{eq:original-problem} is intractable for two reasons: (i) the search space is combinatorial with $2^{N \cdot L}$ possible configurations, and (ii) evaluating accuracy requires a complete forward pass through the model, which is infeasible during inference. We therefore develop a proxy objective that can be optimized efficiently.

\begin{assumption}[Router Score Reliability]
\label{assumption:router-reliability}
The router's gating scores $G_i^{(l)}$ accurately reflects expert importance for token $x_i$ at layer $l$. That is, experts with higher gating scores contribute more to the model's output quality.
\end{assumption}

This assumption is well-supported by prior work. The router is trained end-to-end to select experts that maximize model performance, so high gating scores indicate experts the model deems most relevant. Empirically, \citet{Gupta2024} demonstrates that top-ranked experts contribute the majority of accuracy, while lower-ranked experts provide diminishing returns.

Note that MoE layer was trained in assumption to activate uniform number of experts on each layer, thus we can consider each layer in isolation and following the Assumption \ref{assumption:router-reliability} we can rewrite the optimization problem (\ref{eq:original-problem}) as its per-layer proxy: 

\begin{equation}
\begin{aligned}
\label{eq:proxy-problem}
\max_{S_l \subseteq E} \quad & f_l(S_l; G^{(l)}) = \sum_{j \in S_l} \sum_{i=1}^{n} g_{i,j}^{(l)} \\
\text{s.t.} \quad & |S_l| \leq m_l.
\end{aligned}
\end{equation}

In practice, we use a uniform budget allocation $m_l = K/L$ (described in Section~\ref{sec:practical-algo}), instead of a threshold-based approach  that adaptively determines $|S_l|$ based on the total sum of gating scores at each layer.

\subsection{Modularity and Optimal Greedy Solution}
A key property of the per-layer proxy objective enables efficient optimization:

\begin{proposition}[Modularity of $f_l$]
\label{prop:modularity}
For each layer $l$, the function $f_l(S_l; G^{(l)})$ is modular. That is, for any subset of experts $S \subseteq E$ and expert $e \in E \setminus S$:
\[
f_l(S \cup \{e\}; G^{(l)}) - f_l(S; G^{(l)}) = f_l(\{e\}; G^{(l)}) = \sum_{i=1}^{n} g_{i,e}^{(l)},
\]
independent of $S$.
\end{proposition}
By definition, $f_l(S; G^{(l)}) = \sum_{j \in S} \sum_{i=1}^{n} g_{i,j}^{(l)}$, thus adding expert $e$ contributes exactly $\sum_{i=1}^{n} g_{i,e}^{(l)}$, regardless of which other experts are in $S$.

Modularity implies that the marginal gain from adding any expert is independent of the set of experts that are already included. Consequently, the greedy algorithm, which iteratively adds the expert with maximum marginal gain, produces an \emph{optimal} solution to each per-layer subproblem~\eqref{eq:proxy-problem}.

\begin{corollary}
\label{cor:greedy-optimal}
At each layer $l$, sorting experts by their total sum of gating scores $\sum_{i=1}^n g_{i,j}^{(l)}$ and selecting the top $m_l$ yields an optimal solution to~\eqref{eq:proxy-problem}.
\end{corollary}

Since the main objective~\eqref{eq:original-problem} decomposes across layers and each per-layer subproblem can be solved optimally, the combination of per-layer solutions is optimal for the global proxy problem (given a fixed per-layer budget allocation). Our algorithm requires one additional top-$k$ call, which is negligible in a memory-bound regime.

\subsection{Practical Algorithm}
\label{sec:practical-algo}
Now we introduce a practical implementation of the greedy method which is applied to each layer $l$ independently during inference.

\begin{algorithm}[t]
\caption{Greedy Expert Selection (Per-Layer)}
\label{algo:greedy}
\begin{algorithmic}
\REQUIRE Experts $E$, router scores $G^{(l)}$ for layer $l$,\\ \;\;\;\;\;\;\;\;\;\;\;  threshold $\tau$, initial set $S_0$
\STATE $S \gets S_0$
\\ \STATE\#\# Precompute expert scores for entire batch B
\STATE $\textit{scores}[j] \gets \sum_{i=1}^{n} g_{i,j}^{(l)}$ for all $j \in E$
\WHILE{$f_l(S; P^{(l)}) < \tau \cdot f_l(E; g^{(l)})$} 
    \STATE $e^* \gets \argmax_{e \in E \setminus S} \textit{scores}[e]$
    \STATE $S \gets S \cup \{e^*\}$
\ENDWHILE
\STATE \textbf{return} $S$

\end{algorithmic}
\end{algorithm}

While Corollary~\ref{cor:greedy-optimal} provides an optimal solution to the proxy problem, we find that a variant of Algorithm~\ref {algo:greedy} with a fixed per-layer budget
and warm-up initialization performs better in practice. 

\textbf{Warm-up initialization.} We initialize $S_0$ with the top-1 (or top-$k_0$) experts for each token, ensuring that every token's highest-confidence expert is always included. This guarantees a baseline level of per-token quality before optimizing for batch-level efficiency.

\textbf{Refinement step.} After selecting the expert subset $S_l$ at layer $l$, we route each token to its top-$k$ experts \emph{within} $S_l$. This refinement ensures that the final routing respects the original top-$k$ constraint while restricting choices to the optimized subset.

\noindent During inference, Algorithm~\ref{algo:greedy-full} is applied at each MoE layer $l \in \{1, \ldots, L\}$ as the batch propagates through the network.

\begin{algorithm}[h]
\caption{Batch-Aware Expert Selection (Per-Layer)}
\label{algo:greedy-full}
\begin{algorithmic}
\REQUIRE Experts $E$, batch $B$, router scores $G^{(l)}$,\\ \;\;\;\;\;\;\;\;\;\;\; budget $m_l$, warm-up $k_0$
\STATE \#\# Warm-up: include top-$k_0$ experts per token
\STATE $S_0 \gets \bigcup_{i=1}^{n} \text{Top}\, {k_0}(G_i^{(l)}, E)$ 

\STATE $S_l \gets \textsc{GreedySelect}(E, G^{(l)}, m_l, S_0)$  \# Algorithm~\ref{algo:greedy}
\FOR{each token $x_i \in B$}
    \STATE \#\# Refinement: top-$k$ within selected set
    \STATE Route $x_i$ to $\text{Top}\,k(G_i^{(l)}, S_l)$ 
\ENDFOR
\end{algorithmic}
\end{algorithm}

\section{Speculative Decoding Aware Pruning}
\label{sec:speculative}

Speculative decoding has become a standard technique in modern LLM serving systems to improve throughput. In this approach, a small draft model generates multiple candidate tokens (typically 3--5) that are verified in parallel by the target model. 
While speculative decoding significantly improves token generation rate, it also worsens the expert activation problem: for a batch of $b$ requests with speculative length $L_s$, the effective batch size becomes $b \times (1 + L_s)$. 
For instance, with 8 requests and 3 speculative tokens, the model must process 32 tokens simultaneously, dramatically increasing the number of activated experts. 

\begin{figure}[h]
  \centering
    \centering
    \includegraphics[width=0.6\textwidth]{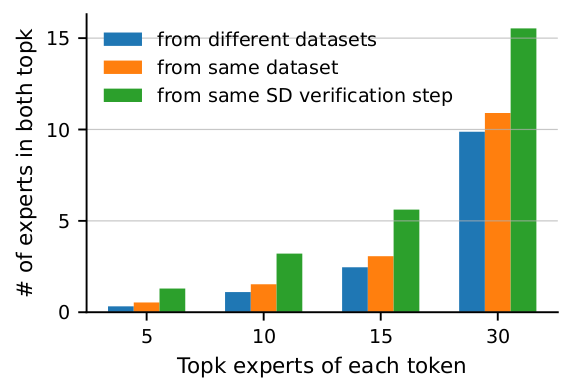}
  \caption{Overlap between the top-$k$ ($k=5,10,15,30$) experts ($|E_i\cap E_j|$) for (1)~two speculative tokens; (2)~two tokens from the same dataset; and (3)~two tokens from different datasets.}
  \label{fig:intersection}
\end{figure}

However, speculative decoding also presents an opportunity: speculative tokens from the same request exhibit strong correlation in their expert preferences, as they represent consecutive steps in the same generation sequence. 
This correlation can be exploited to further reduce expert activation. 
We demonstrate this via the following experiment: we measure the size of the overlap between the top-$k$ selected experts ($|E_i\cap E_j|$) on Figure \ref{fig:intersection}.
As observed, the overlap in expert selection between consecutive speculative tokens is often 2–3× larger than that between two independent tokens drawn from the same dataset. 

\begin{assumption}[Intra-Request Correlation]
\label{assumption:correlation}
Tokens within the same request exhibit correlated expert preferences: if expert $e$ has high gating scores for token $x_r^{(0)}$, it is likely to have high scores for $x_r^{(1)}, \ldots, x_r^{(L_s)}$ as well.
\end{assumption}

\subsection{Hierarchical Proxy for Speculative Decoding}

Consider a batch $B = \{r_1, \ldots, r_b\}$ of $b$ requests, where each request $r$ contains $(1 + L_s)$ tokens: $L_s$ speculative tokens and one generating token. 
Let~$T_r~=~\{x_r^{(0)}, x_r^{(1)}, \ldots, x_r^{(L_s)}\}$ denote the set of tokens belonging to request $r$. The total number of tokens is $n = b \cdot (1 + L_s)$. Similar to Section~\ref{sec:problem-formulation} we aim to solve optimization problem~\ref{eq:original-problem}, given that expert preferences within same request have higher correlation than between requests.

We propose a hierarchical proxy that exploits the correlation structure within requests. For a subset $S$ at layer $l$ and request $r$, define the per-request proxy:
\begin{equation}
f_{l}(S; G^{(l)}; r) = \sum_{x \in T_r} \sum_{j \in S} g_{x,j}^{(l)},
\end{equation}
which measures the total gating score captured by $S$ for all tokens in request $r$. Then, the batch-level proxy decomposes as:
\begin{equation}
\label{eq:speculative-proxy}
f_{\text{spec}}(S; G^{(l)}) = \sum_{r \in B} f_l(S; G^{(l)}, r) = \sum_{r \in B} \sum_{x \in T_r} \sum_{j \in S} g_{x,j}^{(l)}.
\end{equation}

Under Assumption~\ref{assumption:correlation}, optimizing $f_l(S; G^{(l)}; r)$ for each request separately before combining results is more efficient than direct batch-level optimization, as it exploits the shared structure among speculative tokens.
Based on these observations, we propose Algorithm~\ref{algo:spec-request} which performs per-request expert greedy selection and Algorithm~\ref{algo:spec-full} which combines per-request greedy selection with batch-level aggregation.

\begin{algorithm}[t]
\caption{Per-Request Expert Greedy Selection}
\label{algo:spec-request}
\begin{algorithmic}
\REQUIRE Request $r$ with tokens $T_r$, router scores $G^{(l)}$, \\ \;\;\;\;\;\;\;\;\;\;\; per-request budget $m_r$, warm-up $k_0$
\STATE \#\# Warm-up: top-$k_0$ per token
\STATE $S_0 \gets \bigcup_{x \in T_r} \text{Top}\, k_0(G_x^{(l)})$ 
\STATE \#\# Aggregate scores across request
\STATE $\textit{scores}[j] \gets \sum_{x \in T_r} g_{x,j}^{(l)}$ for all $j \in E$ 
\STATE \#\# Add top experts to reach per-request budget
\STATE $S_r \gets S_0 \cup \text{Top}\;{m_r}(\textit{scores}, E \setminus S_0)$ 
\STATE \textbf{return} $S_r$
\end{algorithmic}
\end{algorithm}

\begin{algorithm}[t]
\caption{Speculative Decoding Aware Expert Selection}
\label{algo:spec-full}
\begin{algorithmic}
\REQUIRE Batch $B$, router scores $G^{(l)}$, batch budget $m$, \\ \;\;\;\;\;\;\;\;\;\;\; per-request budget $m_r$, warm-up $k_0$
\FOR{each request $r \in B$}
    \STATE \#\# Calling Algorithm~\ref{algo:spec-request}
    \STATE $S_r \gets \textsc{PerRequestSelect}(r, G^{(l)}, m_r, k_0)$ 
\ENDFOR
\STATE  \#\# Aggregate per-request selections
\STATE $S_{\text{batch}} \gets \bigcup_{r \in B} S_r$
\STATE \#\# Calling Algorithm~\ref{algo:greedy}
\STATE $S_{\text{batch}} \gets \textsc{GreedySelect}(E, G^{(l)}, m_l, S_{\text{batch}})$ 

\FOR{each token $x_i \in B$}
    \STATE \#\# Refinement: top-$k$ within selected set
    \STATE Route $x_i$ to $\text{Top}\, k(G_i^{(l)}, S_{\text{batch}})$ 
\ENDFOR
\STATE \textbf{return} $S_{\text{batch}}$
\end{algorithmic}
\end{algorithm}

The hierarchical approach efficiently exploits the correlation among speculative tokens: since tokens from the same request share context, their expert preferences overlap significantly, allowing a small per-request budget $m_r$ (typically 4--5 experts) to capture most of the gating mass.  

Our experiments in Section~\ref{sec:experiments} show that this approach achieves up to 14\% throughput improvement on GPT-OSS 120B with speculative length 3, while maintaining accuracy within statistical noise of the baseline.

\newpage
\section{Experts Selection for Multi-GPU Inference}
\label{sec:expert-parallelism}

\begin{algorithm}[t]
\caption{GPU-Aware Greedy Expert Selection}
\label{algo:greedy-parallel}
\begin{algorithmic}
\REQUIRE Experts per GPU $E_1, \ldots, E_G$, router scores $G^{(l)}$, \\ \;\;\;\;\;\;\;\;\;\;\; per-GPU budget $m_g$, initial subset $S_0$
\STATE $S \gets S_0$
\WHILE{$|S| < m_g  \cdot G$}
    \FOR{$g \in 1, \ldots, G$}
        \STATE \#\# Best expert on GPU $g$
        \STATE $e^* \gets \argmax_{E_g \setminus S} [f(S\cup\{e\}; G^{(l)}) - f(S; G^{(l)})]$
        
        \STATE $S \gets S \cup \{e^*\}$
    \ENDFOR
\ENDWHILE
\STATE \textbf{return} $S$
\end{algorithmic}
\end{algorithm}

\begin{algorithm}[t]
\caption{Expert Parallelism Aware Selection}
\label{algo:greedy-warm-parallel}
\begin{algorithmic}
\REQUIRE  Experts per GPU $E_1, \ldots, E_G$, router scores $G^{(l)}$,  \\ \;\;\;\;\;\;\;\;\;\;\; batch $B$, per-GPU budget $m_g$, warm-up $k_0$
\STATE \#\# Warm-up: top-$k_0$ per token
\STATE $S_0 \gets \bigcup_{x \in B} \text{Top}_{k_0}(G_x^{(l)})$ 
\STATE \#\# Calling Algorithm~\ref{algo:greedy-parallel}
\STATE $S \gets \textsc{GPUAwareGreedy}(E_1, \ldots, E_G, G^{(l)}, m_g, S_0)$ 
\FOR{each token $x_i \in B$}
    \STATE \#\# Refinement: top-$k$ within selected set
    \STATE Route $x_i$ to $\text{Top}_k(G_i^{(l)}, S)$ 
\ENDFOR
\STATE \textbf{return} $S$
\end{algorithmic}
\end{algorithm}

Expert parallelism (EP) is another promissing technique to scale MoE model training across multiple GPUs. Recently, LLM serving systems such as vLLM and SGLang integrated EP into thier platform to accelerate inference on multiple GPUs. 
In this approach, experts are partitioned across $G$ GPU groups, with each group $g$ hosting a subset of experts $E_g \subseteq E$. 
While expert parallelism enables serving large MoE models that exceed single-GPU memory, it introduces a load balancing challenge: the inference latency for each layer is determined by the GPU with the largest number of activated experts, as all GPUs must synchronize after processing.

However, expert parallelism also presents an optimization opportunity: by considering the GPU placement of experts during selection, we can balance the load across GPUs while maintaining model quality. This is particularly important for production deployments where consistent latency is critical.

\subsection{GPU-Aware Proxy Function}

Consider a system with $G$ GPU groups, where GPU group $g$ hosts experts $E_g$. The experts form a partition: $E~=~\bigcup_{g=1}^{G} E_g$ with $E_g \cap E_{g'} = \emptyset$ for $g \neq g'$. For a batch $B$ with router scores $G^{(l)}$ at layer $l$, we define the per-GPU load for subset $S$:  $\text{Load}_g(S) = |S \cap E_g|$,

which counts the number of selected experts on GPU $g$. The bottleneck load is:
$ \text{MaxLoad}(S) = \max_{g \in \{1,\ldots,G\}} \text{Load}_g(S).$

We maximize our proxy $f(S; G^{(l)})$ while minimizing the bottleneck load, thus GPU-aware optimization problem is:
\begin{equation}
\label{eq:ep-problem}
\max_{S \subseteq E} f(S; G^{(l)}) \quad \text{s.t.} \quad \text{MaxLoad}(S) \leq m_{\text{gpu}},
\end{equation}
where $m_{\text{gpu}}$ is the per-GPU budget.

Standard greedy selection (Algorithm ~\ref{algo:greedy}) can overload a few GPUs with high-scoring experts, while GPU-aware selection trades off gating score to achieve balanced load and lower latency. We modify our greedy approach to select experts per GPU.
At each iteration, instead of selecting the globally best expert, we select the best expert for each GPU group in turn. This ensures that no single GPU accumulates disproportionately many experts. Based on this insight, we propose Algorithm~\ref{algo:greedy-parallel} which performs GPU-balanced greedy selection and Algorithm~\ref{algo:greedy-warm-parallel} which combines warm-up initialization with GPU-aware optimization.

Algorithm~\ref{algo:greedy-parallel} enforces balanced load by design: after $G$ iterations, each GPU has at most one more expert than any other GPU, ensuring $\text{MaxLoad}(S) \leq \lceil |S|/G \rceil$. As previously, it still prioritizes high-scoring experts within each GPU, thus maintaining quality.

\section{Experiments}\label{sec:experiments}

We evaluate all our methods across multiple models (DSR1, GPT-OSS 120B), datasets (AIME2025, GPQA, MMLU-Pro, IFEval, AA-LCR), and deployment scenarios (minimal, speculative decoding, expert parallelism). For details on experimental setup and baselines refer to Appendix~\ref{expsetup}.

\begin{figure*}[h]
    \centering
    \includegraphics[width=1.0\linewidth]{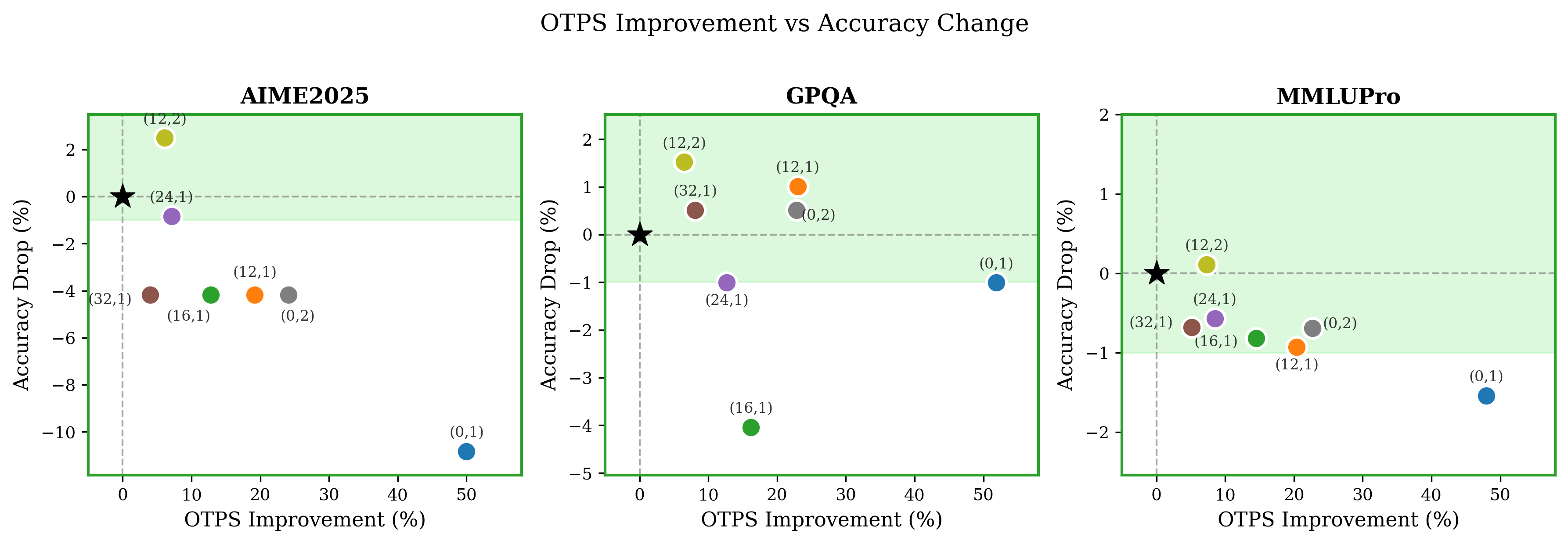}
    \caption{Trade-off between OTPS improvement and accuracy degradation across three datatsets for GPT-OSS 120B (BS=16, speculations off). Each datapoint is accompanied with the pair of parameters defining the algorithm settings (budget $m_l$, warm-up $k_0$).}
    \label{fig:otps-accuracy}
\end{figure*}

\subsection{Minimal Settings - Single GPU, No Speculation}\label{sec:minimalsetting}
We first present experiments in the simplest setting, where the model is hosted on a single GPU and speculative decoding is disabled. We evaluate GPT-OSS-120B using a moderate batch size of 16, which provides a favorable trade-off between latency and cost.  
Figure~\ref{fig:otps-accuracy} shows the trade-off between OTPS improvement and accuracy degradation across three reasoning benchmarks. Two configurations of Algorithm~\ref{algo:greedy-full} consistently achieve throughput gains while keeping accuracy within the 1\% tolerance threshold across all datasets: $(m_l = 24, k_0 = 1)$ delivers 7 -13\% OTPS improvement, and $(12,2)$ provides 6 - 7\% improvement while actually improving accuracy on all three benchmarks. Looking at the Pareto frontier, four configurations $(0,2)$, $(0,1)$, $(12,1)$, and $(12,2)$ consistently represent optimal trade-off points between throughput and accuracy across datasets.

The most aggressive setting $(0,1)$ relies solely on the warm-up set, achieves the best speedup of 50\%, however it comes with a drastic accuracy loss ($-10.83\%$ on AIME2025) . This suggests that warm-up-only expert selection fails to capture the expert utilization patterns required for complex reasoning. Selecting experts that maximize total gating score recovers much of the accuracy loss while retaining meaningful throughput gains. In practice, users should select configurations based on their accuracy tolerance. For full tables with experiments and number of activated experts refer to Appendix~\ref{sec:appendix-experiments}.

\begin{figure*}[h!]
    \centering
    \includegraphics[width=0.9\linewidth]{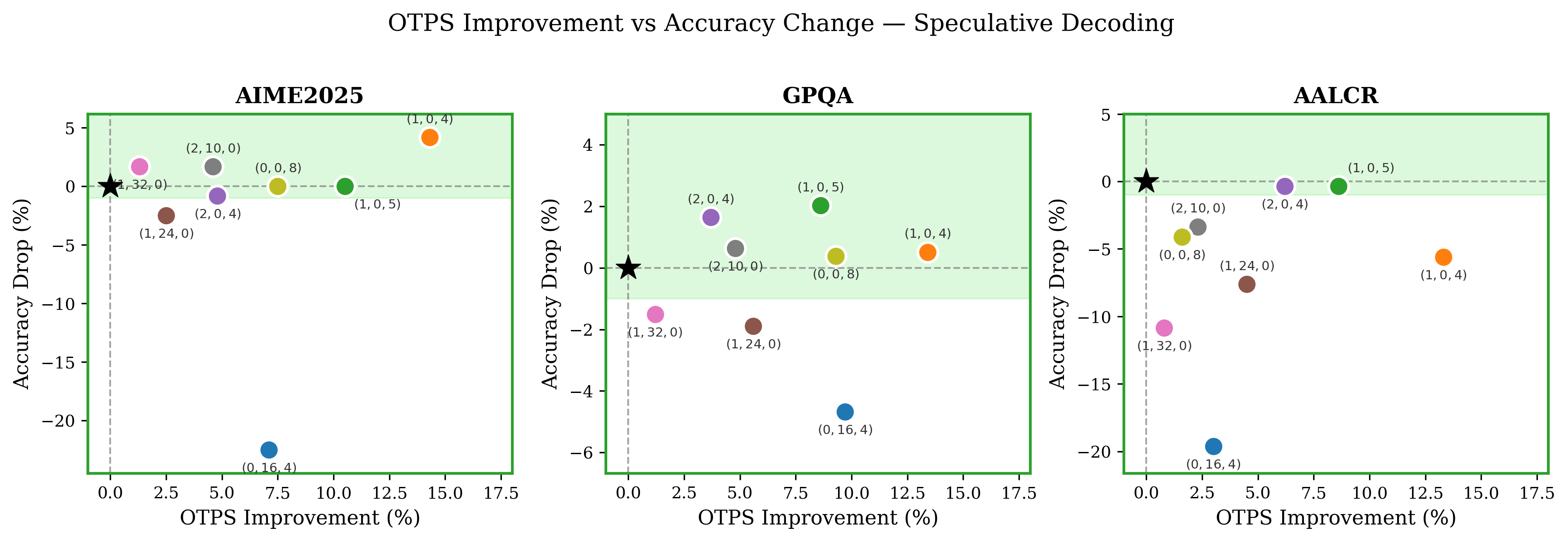}
    \caption{Trade-off between OTPS improvement and accuracy degradation across three datatsets for GPT-OSS 120B (BS=4, speculation length 3). Each datapoint has a set of parameters defining the algorithm settings (warm-up $k_0$, batch budget $m_l$, per-request budget $m_r$).}
    \label{fig:otps-accuracy-spec-dec}
\end{figure*}

\subsection{Speculative Decoding Settings}\label{sec:experiments-speculative}

In Figure~\ref{fig:otps-accuracy-spec-dec} we observe that Algorithm 4, which incorporates speculative decoding-aware expert selection, consistently outperforms Algorithm 2, which does not leverage this structure. 
Algorithm~4 configurations $(k_0 = 1, m = 0,m_r = 4)$ and $(k_0 = 1,m = 0,m_r = 5)$ are Pareto optimal across all three datasets, boosting OTPS by 13--14\% and 8--10\%  respectively while maintaining competitive accuracy. Notably, two configurations consistently preserve accuracy: $(k_0 = 1, m =0,m_r = 5)$ achieves +1.67\%, +2.80\%, and -0.36\% on AIME2025, GPQA, and AALCR respectively, while $(k_0 = 2, m = 0 ,m_r = 4)$ achieves +0.84\%, +2.42\%, and -0.36\% on the same benchmarks.

The configuration $(k_0 = 0, m = 16,m_r = 4)$, is missing warm-up selection and achieves moderate throughput gains ($3-10$\% OTPS) but suffers severe accuracy degradation across all datasets : $-20.83\%$ (AIME2025), $-3.89\%$ (GPQA), $-19.61\%$ (AALCR). This indicates that incorporating even a single token from the warm-up phase is crucial for maintaining accuracy in speculative decoding settings. In contrast, $k_0 \geq 1$ consistently achieves both throughput gains and accuracy preservation. The full tables for this experiments as well as ablation on batch size are provided in Appendix~\ref{sec:appendix-experiments}.

\begin{figure*}[h!]
    \centering
    \includegraphics[width=0.7\linewidth]{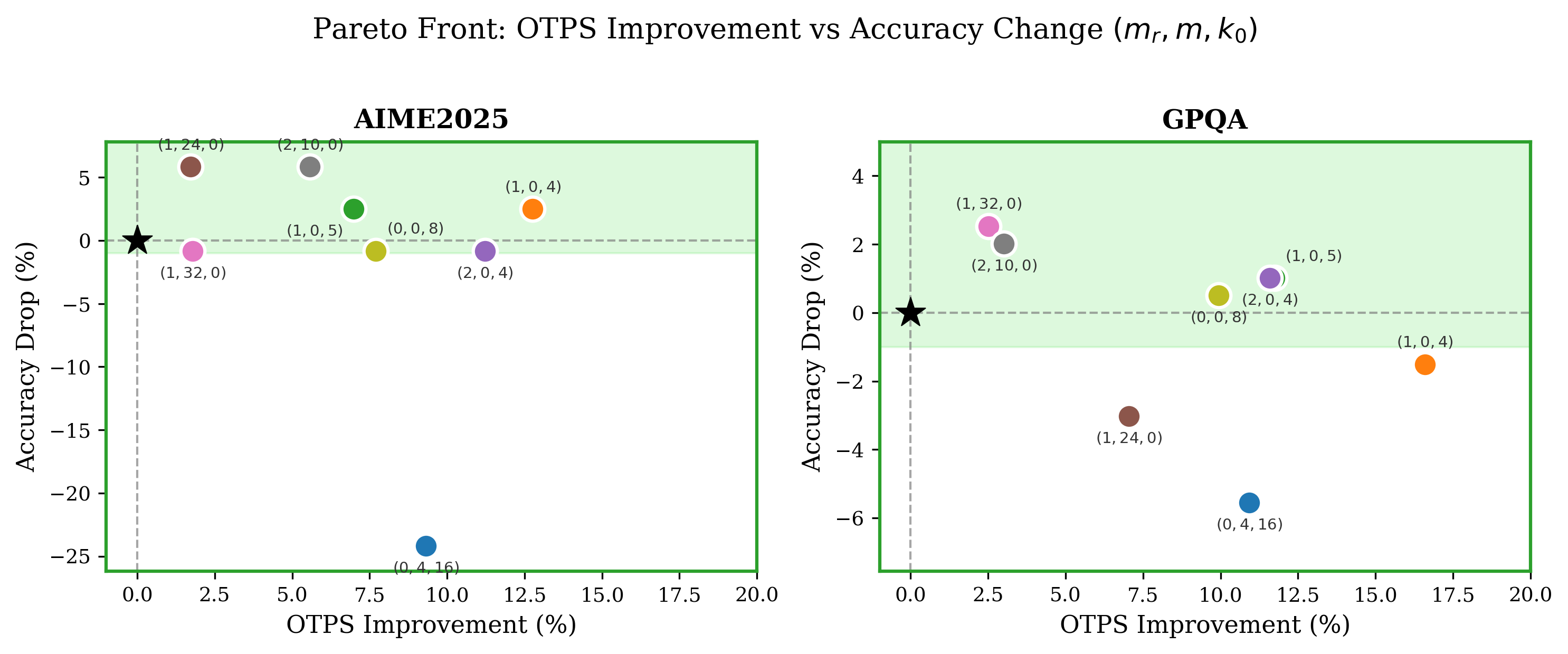}
    \caption{Trade-off between OTPS improvement and accuracy degradation across four datatsets for GPT-OSS 120B (BS=4, speculation length 3) for \textbf{batch with mixed requests}. Each datapoint has a set of parameters defining the algorithm settings (warm-up $k_0$, batch budget $m_l$, per-request budget $m_r$).}
    \label{fig:otps-accuracy-spec-dec-mixed}
\end{figure*}

\setlength{\tabcolsep}{3pt}
\begin{table}[t]
    \centering
    \small
    \begin{tabular}{l|c|cccccccc}
    \toprule
    & Baseline & $(0,4,16)$ & $(1,0, 4)$ & $(1,0, 5)$ & $(2,0, 4)$ & $(1,24, 0)$ & $(1,32, 0)$ & $(2,10, 0)$ & $(0, 0, 8)$ \\
    \midrule
    \multicolumn{10}{c}{\textbf{OTPS}} \\
    \midrule
    AIME2025 & 165.39 & 180.81 & \textbf{186.50} & 176.95 & 183.96 & 168.25 & 168.36 & 174.60 & 178.12 \\
             &        & +9.3\% & \textbf{+12.8\%} & +7.0\% & +11.2\% & +1.7\% & +1.8\% & +5.6\% & +7.7\% \\
    \midrule
    GPQA     & 154.65 & 171.55 & \textbf{180.31} & 172.80 & 172.58 & 165.54 & 158.55 & 159.30 & 170.01 \\
             &        & +10.9\% & \textbf{+16.6\%} & +11.7\% & +11.6\% & +7.0\% & +2.5\% & +3.0\% & +9.9\% \\
    \midrule
    \multicolumn{10}{c}{\textbf{Accuracy}} \\
    \midrule
    AIME2025 & 87.50 & 63.33 & 90.00 & 90.00 & 86.67 & \textbf{93.33} & 86.67 & \textbf{93.33} & 86.67 \\
    Drop     &       & -24.17 & +2.50 & +2.50 & -0.83 & \textbf{+5.83} & -0.83 & \textbf{+5.83} & -0.83 \\
    \midrule
    GPQA     & 76.26 & 70.71 & 74.75 & 77.27 & 77.27 & 73.23 & \textbf{78.79} & 78.28 & 76.77 \\
    Drop     &       & -5.55 & -1.51 & +1.01 & +1.01 & -3.03 & \textbf{+2.53} & +2.02 & +0.51 \\
    \bottomrule
    \end{tabular}
    \vspace{5pt}
    \caption{Trade-off between OTPS improvement and accuracy degradation across four datatsets for GPT-OSS 120B (BS=4, speculation length 3) for \textbf{batch with mixed requests}. Each datapoint is defined by the algorithm settings ( warm-up $k_0$, batch budget $m_l$, request budget $m_r$). For related info refer to Section \ref{sec:experiments-speculative}.}
    \label{tab:otps-accuracy-mixed}
\end{table}

\subsection{Heterogeneous Requests}
In practical deployment, user requests often originate from heterogeneous domains. To reflect this setting in our evaluation, we construct mixed batches consisting of four requests drawn from distinct benchmarks: $r_1$ from GPQA, $r_2$, from AIME2025, $r_3$, from MMLU-Pro, and $r_4$ from AA-LCR. During the verification stage of speculative decoding, each request contributes four tokens, resulting in a batch where requests span diverse datasets, while the speculative tokens within each request remain domain-consistent. We show that this structure enables our method to improve output tokens per second (OTPS) without degrading model quality, even when batches contain requests from multiple, heterogeneous datasets. The main results are reported in Figure~\ref{fig:otps-accuracy-spec-dec-mixed} and Table~\ref{tab:otps-accuracy-mixed}.

\subsection{Expert Parallelism Settings}
\label{epexperiments}
In Table~\ref{tab:deepseek-results} our EP-aware algorithm (Algorithm~\ref{algo:greedy-warm-parallel} with $(k_0 = 1, m_g = 5)$) for DeepSeek-R1 achieves a 73\% drop in total activate experts while maintaining accuracy within 1\% of baseline. It effectively reduces the peak GPU load by 3$\times$ (25.6 $\to$ 8.6 experts), directly translating to improved memory efficiency and reduced communication overhead.

\begin{table}[h!]
    \centering
    \small
    \begin{tabular}{lccc}
    \toprule
    Method & Accuracy & \# Experts & Max/GPU \\
    \midrule
    \multicolumn{4}{c}{\textbf{GSM-8K (Batch Size 8)}} \\
    \midrule
    Original & 0.956 & 50.7 & 11.3 \\
    Algorithm~\ref{algo:greedy-warm-parallel} $(1, 5)$ & 0.946 & 33.2 & 5.94 \\
    \midrule
    \multicolumn{4}{c}{\textbf{IFeval (Batch Size 16)}} \\
    \midrule
    Original & 0.697 & 160.4 & 25.6 \\
    Algorithm~\ref{algo:greedy-warm-parallel} $(1, 5)$ & 0.696 & 43.4 & 8.64 \\
    \bottomrule
    \end{tabular}
    \caption{DS-R1 accuracy and per-GPU load (batch size 8 and 16).}
    \label{tab:deepseek-results}
\end{table}

\section{Conclusion}
We address expert activation explosion in batched MoE inference, where batching and speculative decoding activate 60--95\% of experts despite each token using only $k$.
We introduce the first formal framework for batch-aware expert selection, where maximizing gating mass yields a modular objective with optimal greedy solutions.

We developed 3 algorithms tailored to 3 deployment scenarios: standard serving (batch-aware selection with warm-start initialization), speculative decoding (hierarchical intra-request approach), expert-parallelism (peak GPU load minimizing selection). All algorithms operate at inference time without model retraining and can easily added to existing serving frameworks like vLLM.                       

We practically demonstrate consistent end-to-end throughput improvements measured in output tokens per second, not merely expert sparsity reductions. Algorithm~\ref{algo:spec-full} with $(k_0{=}1, m{=}0, m_r{=}4)$ achieves 13--14\% OTPS improvement across several datasets while maintaining competitive accuracy (GPT-OSS 120B). On DeepSeek-R1 with EP, we cut the number of active experts by 73\%  and cut peak GPU load 3x  while preserving the accuracy.

These results establish that (sub-) modular optimization can achieve practical efficiency gains in production MoE serving without compromising model quality.

\bibliography{references}
\bibliographystyle{abbrvnat}

\newpage
\appendix
\onecolumn

\section{Experimental setup details}\label{expsetup}
\textbf{Models:} We evaluate on two state-of-the-art MoE models: (i) DeepSeek-R1 with 1 shared expert, and 256 routing experts of which 8 are activated per token (\citet{deepseekai2025deepseekr1incentivizingreasoningcapability}), and (ii) GPT-OSS 120B with 128 routing experts of which 4 are activated. (\citet{openai2025gptoss120bgptoss20bmodel}) Both models represent production-scale architectures with different numbers of experts and routing strategies.

\textbf{Hardware and frameworks: }
All inference experiments were conducted on a single p5en.48xlarge EC2 instance equipped with NVIDIA H100 GPUs. We used the vLLM inference engine for all models (\citet{kwon2023efficient}). For GPT-OSS-120B, we ran tensor parallelism with $TP=1$, while for DeepSeek-R1 we used $TP=8$. DeepSeek-R1 was evaluated without speculative decoding. For GPT-OSS-120B, when speculative decoding is reported, we used the EAGLE-3 head with speculation length $3$ (\cite{li2025eagle3}).

\textbf{Datasets:}
We evaluate on a diverse suite of established benchmarks spanning complementary LLM capabilities. AIME2025 (\citet{aime25}) measures advanced mathematical reasoning via competition-style problems requiring multi-step symbolic derivations. GPQA (\citet{rein2024gpqa}) evaluates graduate-level scientific knowledge across physics, chemistry, and biology, with questions designed to be resistant to surface-level pattern matching and to require expert-level reasoning. MMLU-Pro (\citet{wang2024mmlu}) tests broad language understanding across a large collection of multilingual and domain-diverse tasks, emphasizing robust generalization beyond narrow benchmarks. IFEval (\citet{zhou2023instructionfollowingevaluationlargelanguage}) assesses instruction-following fidelity by verifying whether model outputs satisfy explicit natural language constraints. AA-LCR (\citet{aalcr}) targets long-context reasoning, requiring models to integrate and retrieve information across extended inputs with complex dependency structures.

\textbf{Baselines:} We use the original top-$k$ routing of GPT-OSS 120B and DSR1 as baselines. For GPT-OSS 120B, we compare with the baseline on standard batching (Algorithm~\ref{algo:greedy-full}) and speculative decoding scenarios (Algorithm~\ref{algo:spec-full}). For DeepSeek-R1, we compare baseline with GPU-aware expert selection (Algortihm~\ref{algo:greedy-warm-parallel}), since DSR1 has to be served on multiple GPUs.

\section{Additional experiments artifacts}
\label{sec:appendix-experiments}
\subsection {Minimal Settings - Single GPU, No Speculation}
\begin{figure}[h]
    \centering
    \includegraphics[width=0.6\linewidth]{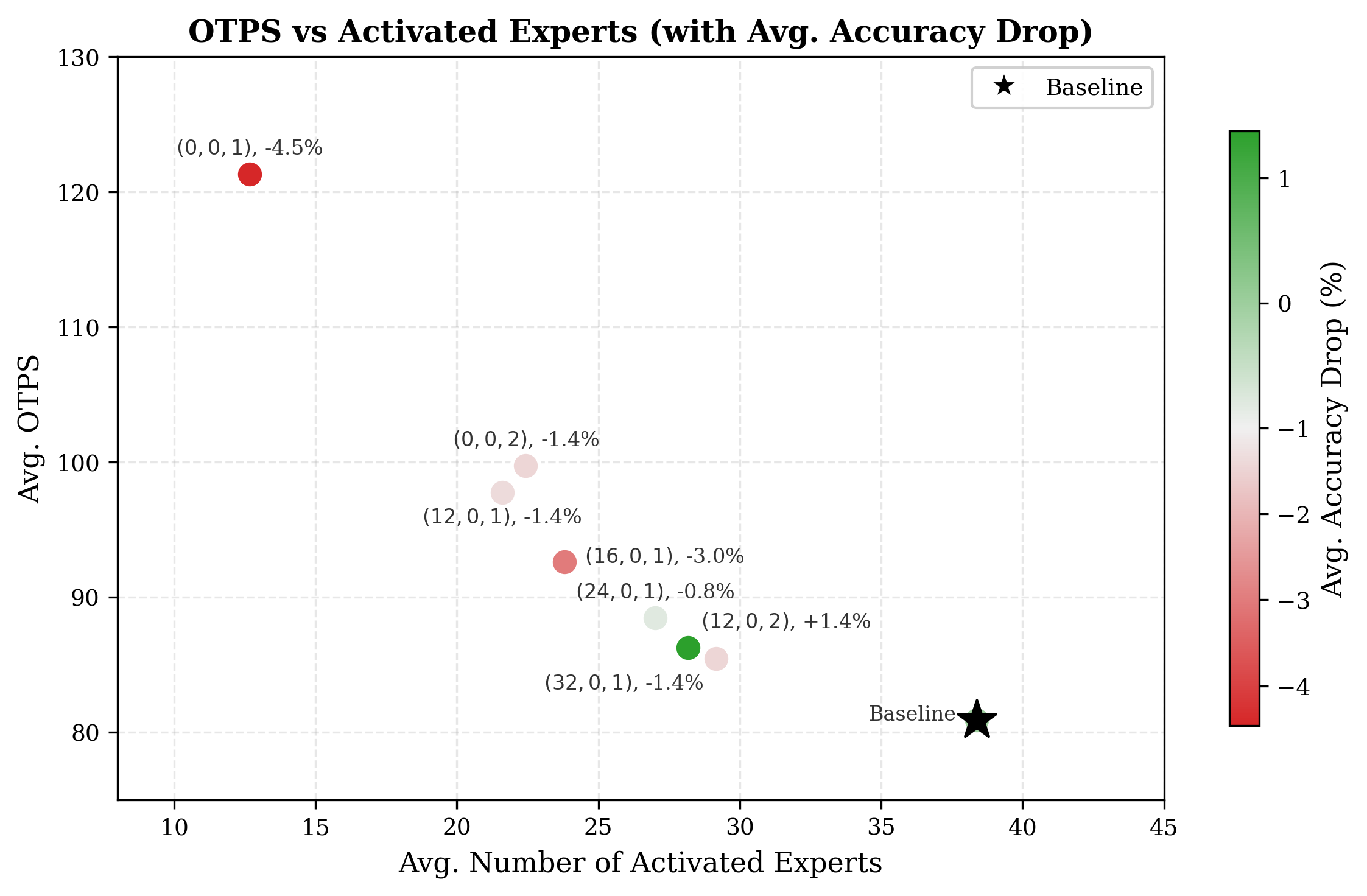}
    \caption{Trade-off between OTPS and \# of activated experts for GPT-OSS 120B (BS=16, speculations off). Each datapoint is defined by the algorithm settings (batch budget $m_l$, request budget $m_r$, warm-up $k_0$). For related info refer to Section \ref{sec:minimalsetting}}
    \label{fig:otps-experts-no-sd}
\end{figure}

\begin{itemize}
    \item For all the data points used in Section \ref{sec:minimalsetting} refer to table \ref{tab:gptoss-throughput-no-spec}.
    \item For tradeoff on activated experts vs. OTTPS refer to table Figure \ref{fig:otps-experts-no-sd}
\end{itemize}
\begin{table}[h]
    \centering
    \small
    \begin{tabular}{l|c|cccccccc}
    \toprule
    & Baseline & $(0,1)$ & $(12,1)$ & $(16,1)$ & $(24,1)$ & $(32,1)$ & $(0,2)$ & $(12,2)$ & $(24,0)$ \\
    \midrule
    \multicolumn{10}{c}{\textbf{OTPS}} \\
    \midrule
    AIME2025 & 85.83 & \textbf{128.72} & 102.35 & 96.85 & 91.97 & 89.23 & 106.49 & 91.04 & 107.09 \\
             &       & \textbf{+50.0\%} & +19.2\% & +12.8\% & +7.1\% & +4.0\% & +24.1\% & +6.1\% & +24.8\% \\
    \midrule
    GPQA     & 75.48 & \textbf{114.65} & 92.81 & 87.70 & 85.05 & 81.60 & 92.71 & 80.36 & 102.49 \\
             &       & \textbf{+51.9\%} & +23.0\% & +16.2\% & +12.7\% & +8.1\% & +22.8\% & +6.5\% & +35.8\% \\
    \midrule
    MMLUPro  & 81.40 & \textbf{120.49} & 98.03 & 93.21 & 88.33 & 85.53 & 99.88 & 87.37 & 104.84 \\
             &       & \textbf{+48.0\%} & +20.4\% & +14.5\% & +8.5\% & +5.1\% & +22.7\% & +7.3\% & +28.8\% \\
    \midrule
    \multicolumn{10}{c}{\textbf{Accuracy}} \\
    \midrule
    AIME2025 & 87.50 & 76.67 & 83.33 & 83.33 & 86.67 & 83.33 & 83.33 & \textbf{90.00} & - \\
    Drop     &       & -10.83 & -4.17 & -4.17 & -0.83 & -4.17 & -4.17 & \textbf{+2.50} & - \\
    \midrule
    GPQA     & 76.26 & 75.25 & 77.27 & 72.22 & 75.25 & 76.77 & 76.77 & \textbf{77.78} & - \\
    Drop     &       & -1.01 & +1.01 & -4.04 & -1.01 & +0.51 & +0.51 & \textbf{+1.52} & - \\
    \midrule
    MMLUPro  & 80.69 & 79.15 & 79.76 & 79.87 & 80.12 & 80.01 & 80.00 & \textbf{80.80} & - \\
    Drop     &       & -1.54 & -0.93 & -0.82 & -0.57 & -0.68 & -0.69 & \textbf{+0.11} & - \\
    \bottomrule
    \end{tabular}
    \caption{Trade-off between OTPS improvement and accuracy degradation across three datatsets for GPT-OSS 120B (BS=16, speculations off). Each datapoint is defined by the algorithm settings (budget ml, warm-up k0). For related info refer to Section \ref{sec:minimalsetting}}
    \label{tab:gptoss-throughput-no-spec}
\end{table}

\subsection{Speculative Decoding Settings}

\begin{figure}[h]
    \centering
    \includegraphics[width=0.6\linewidth]{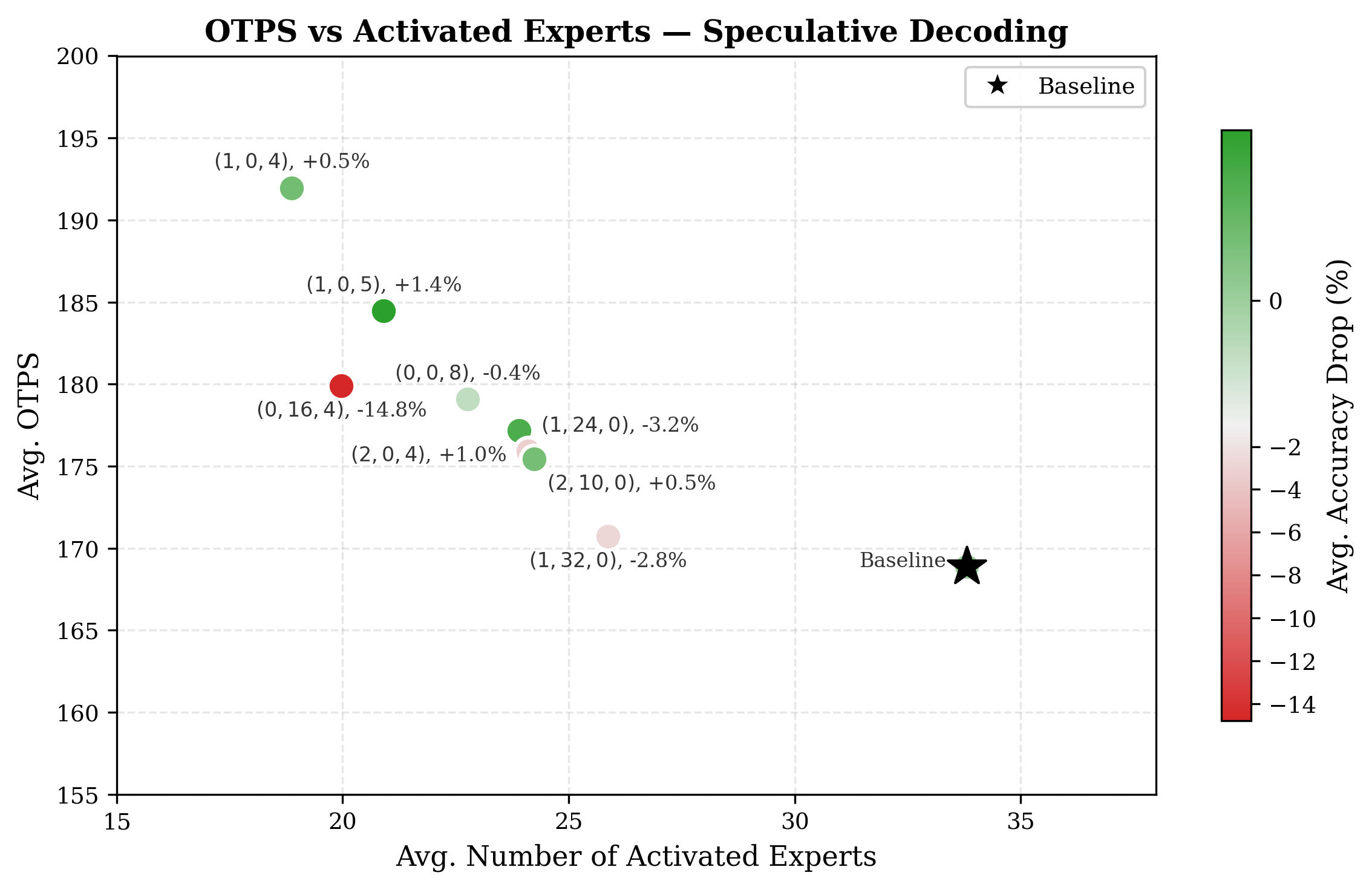}
    \caption{Trade-off between OTPS and \# of activated experts for GPT-OSS 120B (BS=4, speculation length 3). Each datapoint is defined by the algorithm settings (warm-up $k_0$, batch budget $m_l$, request budget $m_r$ ). For related info refer to Section \ref{sec:experiments-speculative}}
    \label{fig:otps-experts-sd}
\end{figure}

\begin{itemize}
    \item For all the data points used in Section \ref{sec:experiments-speculative} refer to table \ref{tab:otps-accuracy}.
    \item For tradeoff on activated experts vs. OTTPS refer to table Figure \ref{fig:otps-experts-sd}
\end{itemize}

\setlength{\tabcolsep}{3pt}
\begin{table*}[h]
    \centering
    \small
    \begin{tabular}{l|c|cccccccc}
    \toprule
    & Baseline & $(0,16,4)$ & $(1,0,4)$ & $(1,0,5)$ & $(2,0,4)$ & $(1,24,0)$ & $(1,32,0)$ & $(2,10,0)$ & $(0,0,8)$ \\
    \midrule
    \multicolumn{10}{c}{\textbf{OTPS}} \\
    \midrule
    AIME2025 & 173.07 & 185.38 & \textbf{197.78} & 191.23 & 181.35 & 177.41 & 175.28 & 181.10 & 186.06 \\
             &        & +7.1\% & \textbf{+14.3\%} & +10.5\% & +4.8\% & +2.5\% & +1.3\% & +4.6\% & +7.5\% \\
    \midrule
    IFBench  & 163.82 & 180.60 & \textbf{191.51} & 178.91 & 171.15 & 172.67 & 170.21 & 170.32 & 178.04 \\
             &        & +10.2\% & \textbf{+16.9\%} & +9.2\% & +4.5\% & +5.4\% & +3.9\% & +4.0\% & +8.7\% \\
    \midrule
    LCBench  & 174.01 & 179.23 & \textbf{197.19} & 188.96 & 184.75 & 181.80 & 175.37 & 177.97 & 176.72 \\
             &        & +3.0\% & \textbf{+13.3\%} & +8.6\% & +6.2\% & +4.5\% & +0.8\% & +2.3\% & +1.6\% \\
    \midrule
    MMLUPro  & 170.07 & 182.06 & \textbf{195.34} & 188.03 & 177.05 & 177.59 & 174.40 & 183.73 & 177.05 \\
             &        & +7.1\% & \textbf{+14.9\%} & +10.6\% & +4.1\% & +4.4\% & +2.5\% & +8.0\% & +4.1\% \\
    \midrule
    GPQA     & 159.52 & 175.06 & \textbf{180.91} & 173.20 & 165.44 & 168.50 & 161.47 & 167.24 & 174.48 \\
             &        & +9.7\% & \textbf{+13.4\%} & +8.6\% & +3.7\% & +5.6\% & +1.2\% & +4.8\% & +9.3\% \\
    \midrule
    \multicolumn{10}{c}{\textbf{Accuracy}} \\
    \midrule
    AIME2025 & 87.50 & 65.00 & \textbf{91.67} & 87.50 & 86.67 & 85.00 & 89.17 & 89.17 & 87.50 \\
    Drop     &        & -22.50 & \textbf{+4.17} & 0.00 & -0.83 & -2.50 & +1.67 & +1.67 & 0.00 \\
    \midrule
    GPQA     & 76.26 & 71.59 & 76.77 & \textbf{78.28} & 77.90 & 74.37 & 74.75 & 76.89 & 76.64 \\
    Drop     &        & -4.67 & +0.51 & \textbf{+2.02} & +1.64 & -1.89 & -1.51 & +0.63 & +0.38 \\
    \midrule
    AALCR    & 48.86 & 29.25 & 43.25 & \textbf{48.50} & \textbf{48.50} & 41.25 & 38.00 & 45.50 & 44.75 \\
    Drop     &        & -19.61 & -5.61 & \textbf{-0.36} & \textbf{-0.36} & -7.61 & -10.86 & -3.36 & -4.11 \\
    \bottomrule
    \end{tabular}
    \caption{Trade-off between OTPS improvement and accuracy degradation across three datatsets for GPT-OSS 120B (BS=4, speculation length 3). Each datapoint is defined by the algorithm settings ( warm-up $k_0$, batch budget $m_l$, request budget $m_r$). For related info refer to Section \ref{sec:experiments-speculative}.}
    \label{tab:otps-accuracy}
\end{table*}

\end{document}